\documentclass[9pt,conference]{IEEEtran}
\IEEEoverridecommandlockouts
\usepackage{amsmath,amssymb,subcaption,booktabs, multirow,amsthm}

\usepackage{setspace}

\usepackage[utf8]{inputenc} 
\usepackage[T1]{fontenc}    
\usepackage{hyperref}       
\usepackage{url}            
\usepackage{booktabs}       
\usepackage{amsfonts}       
\usepackage{nicefrac}       
\usepackage{microtype}      
\usepackage{xcolor}         

\usepackage{algorithm}
\usepackage[noend]{algorithmic}
\usepackage{caption}
\usepackage{enumitem}
\usepackage[pdftex]{graphicx}
\usepackage{wrapfig}
\usepackage{multirow}

\usepackage{xr}

\DeclareMathOperator*{\argmax}{arg\,max}

\title{Peer-to-Peer Deep Learning for Beyond-5G IoT}
\author{\IEEEauthorblockN{Srinivasa Pranav}
\IEEEauthorblockA{\textit{Electrical and Computer Engineering} \\ \textit{Carnegie Mellon University } \\ Pittsburgh, USA \\
spranav@cmu.edu}
\and
\IEEEauthorblockN{Jos\'e M.F. Moura}
\IEEEauthorblockA{\textit{Electrical and Computer Engineering} \\ \textit{Carnegie Mellon University } \\ Pittsburgh, USA \\
moura@andrew.cmu.edu}
\thanks{
This work was partially supported by NSF Grant CCF-2327905 and XSEDE~\cite{6866038}
Allocation ELE220003
on PSC Bridges-2~\cite{10.1145/3437359.3465593}. First author partially supported by a NSF Graduate Research Fellowship (GRFP; Grants DGE1745016, DGE2140739), and an ARCS Fellowship. 
}
}

\begin{document}
\maketitle
\begin{abstract}
We present P2PL, a practical multi-device peer-to-peer deep learning algorithm that, unlike the federated learning paradigm, does not require coordination from edge servers or the cloud. This makes P2PL well-suited for the sheer scale of beyond-5G computing environments like smart cities that otherwise create range, latency, bandwidth, and single point of failure issues for federated approaches. 

P2PL introduces max norm synchronization to catalyze training, retains on-device deep model training to preserve privacy, and leverages local inter-device communication to implement distributed consensus. Each device iteratively alternates between two phases: 1) on-device learning and 2) peer-to-peer cooperation where they combine model parameters with nearby devices. We empirically show that \textit{all} participating devices achieve the same test performance attained by federated and centralized training -- even with 100 devices and relaxed singly stochastic consensus weights. We extend these experimental results to settings with diverse network topologies, sparse and intermittent communication, and non-IID data distributions.
\end{abstract}
\section{Introduction}
\label{sec:intro}
Beyond-5G internet-of-things (IoT) is evolving into vast networks of mobile devices that wirelessly collaborate with their peers to train intelligent models and perceive the world around them. IoT, smart cities, and other emerging computing environments are densely populated with devices that share three key characteristics:
1) they autonomously collect and process data on-device; 2) they cooperate with \textit{peers} over wireless device-to-device links; 3) they have great potential to leverage advancements in \textit{deep learning}.

Innovative applications for emerging distributed systems, like autonomous vehicles, rely on deep neural networks to achieve very high performance on tasks like image classification~\cite{9706268}. Simultaneously, hardware developments are enabling deep neural network training on mobile devices~\cite{10.1145/3486618, patil2022poet} and inference on ultra-low-power embedded devices~\cite{10.1145/3510850}. Distributed devices in beyond-5G IoT can not solely rely on local training data since it is often insufficient for training deep models that generalize well to unseen data~\cite{pmlr-v54-mcmahan17a}. As these networked systems are rapidly growing, scaling privacy-preserving cloud and edge server-based collaboration among devices (i.e., federated learning~\cite{pmlr-v54-mcmahan17a}) is increasingly infeasible. This makes it crucial for devices to cooperate with nearby peers over wireless device-to-device links.

Peer-to-peer Learning (P2PL), our device-to-device deep learning framework, enables distributed IoT devices to collaboratively train deep models and achieve high test performance without relying on coordinating servers. We demonstrate P2PL's robustness to independent and random initialization, diverse physical arrangements of devices, temporal communication sparsity, and non-IID data distribution.

\textbf{Contributions}: We present P2PL, a device-to-device deep learning framework that involves a learning phase with on-device training of deep models and a consensus phase with inter-device communication. We empirically show that our framework enables large connected networks of devices (without edge or cloud servers) to collaboratively learn deep models that achieve the same test performance as federated and centralized learning frameworks. Our experiments demonstrate P2PL's max norm synchronization phase producing faster convergence than existing device-to-device methods. Our results encourage relaxing doubly stochastic mixing weight assumptions used in theoretical analysis. We empirically show that P2PL is resilient to environments with link failures, sparse gossip communication, and non-IID data. 

\section{Related Work}
\label{sec:previous_work}
\subsection{Centralized and Federated Deep Learning} \label{sec:centralized_and_federated}
Early \textit{centralized} approaches relied on massive datacenters to aggregate data from devices like computers and mobile phones. Training deep neural networks on data split across multiple nodes required co-optimizing datacenter architectures and training algorithms~\cite{dean2012large, moritz2018ray}.
However, privacy concerns and the EU GDPR legislation~\cite{GDPR16} motivated privacy-preserving edge computing approaches.

\begin{figure}[t]
\centering
\includegraphics[width=0.5\columnwidth]{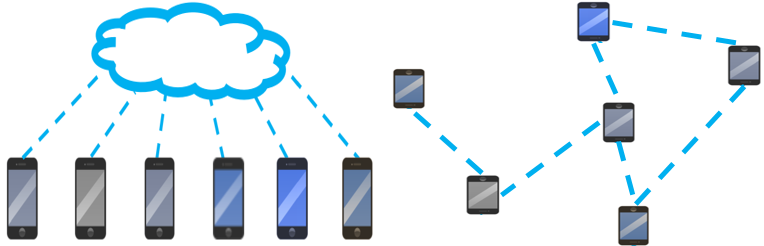}
\caption{Federated setting (left); Peer-to-Peer setting (right).}
\label{fig:1}
\end{figure}

\textit{Federated} deep learning approaches~\cite{pmlr-v54-mcmahan17a} involve devices locally training deep models on raw data that never leaves the device. Devices instead transmit trained model parameters to the cloud (Fig.~\ref{fig:1}, left) or to servers at the network edge~\cite{hierarchical}. Servers act as fusion centers by combining local model parameters to create a global model and broadcasting it back to the devices.
The server plays a essential part in almost all state-of-the-art techniques for training deep models across devices~\cite{9084352} and it limits those techniques to federated settings. As the number of devices grows in beyond-5G environments, the federated computing model faces many issues: range limitations, high latency, bandwidth bottlenecks, and single point of failure.

\subsection{Peer-to-Peer Deep Learning}
\label{section:p2pdl}
\textit{Peer-to-peer} beyond-5G infrastructureless systems (Fig.~\ref{fig:1}, right) leverage local device-to-device communication to bypass the issues faced by federated approaches. Peer-to-peer learning enables deploying vast systems, reduces latency, avoids bottlenecks, removes the single point of failure, and can reduce costs. These reasons motivated significant work in distributed optimization and distributed signal processing over the past 15 years.
This paper's framework is based on Consensus+Innovations~\cite{kar2013consensus+, kar2012distributed}, a device-to-device algorithm that alternates between a consensus step, where devices exchange their state with nearby devices, and an innovations step, where devices process local data and received state information. Other peer-to-peer distributed optimization algorithms, like Distributed Nesterov methods~\cite{6705625, 8812696}, focused on (strongly) convex objective functions.
However, the vast representation power and empirical success of deep neural networks justify shifting our focus to highly nonconvex problems.

More recent works~\cite{jiang2017collaborative, lian2017can, cooperativeSGD} analyzed peer-to-peer variants of stochastic gradient descent (SGD) in nonconvex settings (under additional assumptions) and found convergence bounds on the expected gradient norm that have asymptotic error floors. Refs.~\cite{pmlr-v119-koloskova20a, li2021communicationefficient} derived similar bounds for local SGD (devices perform multiple local gradient updates between consensus steps), but all of these works assume that consensus steps use doubly stochastic mixing weights: each pair of devices assigns the same importance to each other's parameters. This conflicts with the intuition that devices with more local training data will produce models that generalize better to unseen data~\cite{pmlr-v54-mcmahan17a}. We empirically show that using singly stochastic mixing weights, based on local dataset sizes, leads to faster convergence than the popular doubly stochastic Metropolis-Hastings mixing weights~\cite{gossip-metropolis-hastings-mixing}.
To evaluate algorithm performance in much larger and more diverse settings than in~\cite{jiang2017collaborative, lian2017can, cooperativeSGD, pmlr-v119-koloskova20a, li2021communicationefficient}, our experiments use 100 devices and a diverse communication graphs.

Consensus-based federated averaging (CFA)~\cite{CFA} also uses an iterative two-step process of on-device training and peer-to-peer consensus, but there are notable differences with our work: 1) CFA starts with a consensus step that pulls model parameters closer to the origin; 2) CFA does not use device-specific consensus step sizes and the default mixing weights ignore the receiving device's dataset size; 3) CFA does not use momentum for on-device training. Our experiments show that the former two design choices deteriorate performance, slowing the convergence rate. Our approach, P2PL, is distinct from CFA and addresses these issues by starting with a maximum norm synchronization phase, allowing device-specific consensus step sizes, and using local momentum.

\section{Peer-to-Peer Deep Learning} \label{sec:details}
\subsection{Optimization Problem Setup}
We consider $K$ devices with independently and randomly initialized instances of a parametric machine learning model (i.e., deep neural network) with corresponding parameters $w_1,\dots,w_K$.
Let $\mathcal{D} = \{(x_1, y_1),\dots,(x_n, y_n)\}$ denote the full training dataset at the global level. We view devices' local datasets $\mathcal{D}_1,\dots,\mathcal{D}_K$ as the partitioning of $\mathcal{D}$ across $K$ devices. Let $n_k=|\mathcal{D}_k|$ be the $k$th device's local training dataset size. The partition reflects each device only accessing its unique local dataset that never leaves the device. 

Let $\ell_w(x,y)$ be the loss incurred by a model, parameterized by $w$, when evaluated on sample $(x,y)$. In \eqref{eq:1}, let $F_k(w)$ be the average empirical loss of a model, parameterized by $w$, when evaluated on $\mathcal{D}_k$.
In \eqref{eq:objective}, we formulate peer-to-peer deep learning as the same weighted finite-sum optimization objective as federated learning~\cite{pmlr-v54-mcmahan17a}.

\begin{equation}
F_k(w) = \frac{1}{n_k}\sum_{(x_i,y_i)\in\mathcal{D}_k} \ell_w(x_i,y_i) \label{eq:1}
\end{equation}

\begin{equation}
    \min_{w} \frac{1}{n} \sum_{k=1}^K n_k F_k(w)
    \label{eq:objective}
\end{equation} 

We want all devices' models to converge to the same local minimum. Letting $w_k^{(t)}$ be the $k$th device's model parameters at round $t$ of an iterative optimization algorithm, global consensus is defined by \eqref{eq:3}:
\begin{equation}
    \forall k\in[K],\;\; \lim_{t\to\infty} w_k^{(t)} = w^{(\infty)}
    \label{eq:3}
\end{equation}
Consensus, along with limited data at each device in practice, makes collaboration between devices essential for achieving high performance on test data. Since peer-to-peer settings lack server-facilitated collaboration, we leverage device-to-device communication.
\subsection{Device-to-Device Communication and Connectivity}
\label{sec:comms}
This paper focuses on discrete-time optimization and consensus algorithms that proceed in rounds. We assume device-to-device communication is bidirectional (full-duplex), synchronous, and noiseless. Unlike centralized and federated settings, the physical configuration of devices significantly affects convergence in peer-to-peer settings.

Device capabilities and the physical environment induce a network topology, where direct communication is only feasible between certain pairs of devices. This is modeled by a communication graph with vertices representing devices and edges indicating direct wireless channels between nearby devices. Let $\mathcal{N}(k)$ denote the set of device $k$'s neighbors. We consider network topologies modeled by fixed, flat, simple, undirected, and connected communication graphs. These include random graphs like 3D random geometric graphs that realistically model ad hoc mobile device network topologies~\cite{RGG}. In contrast, federated learning implicitly considers a star graph with a server at the center.\footnote{In federated learning, the server acts as a fusion center and does not have its own training data. It only aggregates parameters and rebroadcasts to devices.} 

\subsection{Peer-to-Peer Learning Algorithm}
Peer-to-Peer Learning (P2PL) is an iterative fully distributed optimization algorithm with a preliminary maximum norm synchronization phase followed by update rounds split into training and consensus phases. Pseudocode is given in Algorithm~\ref{alg:P2PL}.

The max norm synchronization phase ensures that future consensus steps correspond to averaging relevant accumulated gradient updates from each device's neighbors. Each device exchanges its deep model parameters with its neighbors $\mathcal{N}(\cdot)$ and updates its own parameters using \eqref{eq:max_norm_sync}. All devices' parameters are synchronized when this process repeats $\textrm{diameter}(G)$ times (where~\cite{peleg2012distributed-diameter} can compute the diameter).
\begin{equation}
    \label{eq:max_norm_sync}
    w_k \leftarrow w_j : j = \argmax_{i \in \mathcal{N}(k) \cup \{k\}} \|w_i\|_F
\end{equation}

During the training phase, each device trains its deep model for one epoch with local training data and mini-batch gradient descent (GD) with momentum. Pseudocode in Algorithm~\ref{alg:P2PL} uses $v_k$ in lines 10-11 to describe the default PyTorch~\cite{Pytorch_NEURIPS2019_9015} implementation of momentum (a variant of Polyak momentum~\cite{polyak1964some}).

For the consensus phase, each device exchanges model parameters with its neighbors $\mathcal{N}(\cdot)$ and updates its own parameters with the weighted convex combination in \eqref{eq:consensus}. This provides a communication-efficient method for updating a device's parameters with gradient information obtained by neighbors training on their local datasets. 
\eqref{eq:consensus} fully distributes the server's aggregation step in federated learning~\cite{pmlr-v54-mcmahan17a} and makes each device act like a server in its vicinity. 
\begin{equation}
    w_k \leftarrow w_k + \epsilon^{(t)}_k \sum_{i\in\mathcal{N}(k)}\alpha_{ki}(w_k - w_i)
    \label{eq:consensus}
\end{equation}
$\epsilon^{(t)}_k$ is a time-varying, device-specific consensus step size (a tunable hyperparameter). $\alpha_{ki}$ is the mixing weight that device $k$ gives to device $i$'s parameters. Motivated by how deep models trained with more data tend to generalize better to unseen data, we can choose
\begin{equation}
\label{eq:mixing-weights}
    \forall k,i\in[K],\;\;\alpha_{ki} = \frac{n_i}{n_k + \sum_{j \in \mathcal{N}(k)}n_j}
\end{equation}

\begin{algorithm}[t]
\footnotesize
\caption{\fontsize{9}{10}\selectfont
Peer-to-Peer Learning (P2PL)\\ \textit{B}, batch size; \textit{$\eta$}, learning rate; \textit{$\mu$}, momentum; $\mathcal{D}_k$, \textit{k}th local dataset.}
\label{alg:P2PL}
\begin{algorithmic}[1]
\STATE 
\textbf{Each device \textit{k} executes the following:}
\STATE $\mathcal{B}_k \leftarrow$ (split $\mathcal{D}_k$ into batches of size \textit{B}), $v_k\leftarrow 0$, and initialize $w_k$
\STATE\COMMENT{Max Norm Synchronization Phase:}
\FOR{$t = 1$ to Diameter(G)}
\STATE Exchange $w_k$ with neighbors $i\in\mathcal{N}(k)$
\STATE $w_k \leftarrow w_j : j = \argmax_{i \in \mathcal{N} \cup \{k\}} \|w_i\|_F$
\ENDFOR
\FOR{each round $t = 1,2,\dots$}
\STATE\COMMENT{Training Phase:}
\FOR{each batch $b \in \mathcal{B}_k$}
\STATE $v_k \leftarrow \mu v_k + \frac{1}{B}\sum_{(x_i,y_i)\in b}\nabla l_{w_k}(x_i,y_i)$
\STATE $w_k \leftarrow w_k - \eta v_k$
\ENDFOR
\STATE\COMMENT{Consensus Phase:}
\STATE Exchange $(n_k,w_k)$ with neighbors $i\in\mathcal{N}(k)$
\STATE $w_k \leftarrow w_k + \epsilon^{(t)}_k \sum_{i\in\mathcal{N}(k)} \alpha_{ki}(w_k - w_i)$
\ENDFOR
\end{algorithmic}
\end{algorithm}

\section{Experimental Results and Discussion}
\subsection{Experimental Setup} \label{sec:setup}
We simulate P2PL on $K{=}100$ devices (significantly more than prior works~\cite{jiang2017collaborative, lian2017can, CFA, cooperativeSGD}). \textbf{Dataset}: MNIST handwritten digit classification~\cite{lecun2010mnist}, where each $x_i$ is a $28\times28$ normalized single-channel image depicting a handwritten label $y_i \in \{0,1,\dots,9\}$. Our data is partitioned in a similar manner as Ref.~\cite{pmlr-v54-mcmahan17a}: for the \textbf{IID} setting, the 60,000 training samples are shuffled and partitioned equally into 100 local device training datasets, ($\forall k\; n_k$ = 600); for the balanced, pathological \textbf{Non-IID} setting, the 60,000 training samples are sorted by label, partitioned into 200 shards, and each of the $K$ = 100 distributed devices is randomly assigned 2 shards. \textbf{Architecture:} each device uses PyTorch's default independent and random initialization~\cite{Pytorch_NEURIPS2019_9015} for a 2NN: a two-layer perceptron architecture with ReLU activation and 200 nodes in each hidden fully-connected layer. \textbf{Optimization}: mini-batch gradient descent with the PyTorch~\cite{Pytorch_NEURIPS2019_9015} default variant of Polyak momentum (see Alg.~\ref{alg:P2PL}), where $B$ = 10, fixed $\eta$ = 0.01, $\mu$ = 0.5, $\forall t,k\; \epsilon_k^{(t)}$ = 1, and $\alpha_{ki}$ are set according to \eqref{eq:mixing-weights}. 

\textbf{Test Metric}: test accuracy achieved by devices on 10,000 MNIST test samples.
\textbf{Convergence}: the first round at which local deep models at \textbf{all} $K$ = $100$ devices cross 97\% test accuracy (the threshold used by Ref.~\cite{pmlr-v54-mcmahan17a}). This definition of convergence tracks the worst-performing device and is \textit{very conservative} since, in practice, we are usually concerned with average rather than worst case analysis. However, it offers the fairest comparison with baselines. \textbf{Baselines}: the centralized setting uses single-model momentum mini-batch gradient descent; the federated setting uses Federated Averaging (FedAvg)~\cite{pmlr-v54-mcmahan17a} with full participation and momentum mini-batch gradient descent at each device. Our communication budget is 10,000 communication rounds.

\subsection{Insights from a Complete Communication Graph}
\label{sec:complete_graph}

We demonstrate that P2PL leverages max norm synchronization to converge up to 3x faster than CFA~\cite{CFA}. These experiments use a complete communication graph (all pairs of devices are neighbors) to allow all devices to replicate the behavior and convergence rate of a server in federated settings (with full participation). In both federated and peer-to-peer settings, consensus phases are designed for devices to incorporate relevant accumulated gradient updates from their neighbors. This intended effect is achieved when all devices start with identical model parameters before training~\cite{pmlr-v54-mcmahan17a}. In fully distributed settings, the lack of a coordinating server naturally leads to devices independently and randomly initializing their deep models.

\begin{table}[t]
\footnotesize
\centering
\caption{\fontsize{9}{10}\selectfont Rounds required for minimum test accuracy of $K{=}100$ independently, randomly initialized devices to cross $97\%$. Distributed trials (last 4) use a complete communication graph. *Our methods.}
\begin{tabular}{lc}
\toprule
 Algorithm & Rounds to Convergence \\
\midrule
FedAvg~\cite{pmlr-v54-mcmahan17a} &  \textbf{96}\\
P2PL* &  \textbf{96}\\
P2PL without Max Norm Synchronization & 155\\
CFA~\cite{CFA} with Momentum & 168\\
CFA~\cite{CFA} & 294\\
\bottomrule
\end{tabular}
\label{table:fed_vs_dist}
\end{table}

Immediately iterating training and consensus phases from independent and random initialization corresponds to P2PL without the max norm synchronization phase. In early rounds, many devices are forced to incorporate irrelevant gradient information from neighbors with greatly differing model parameters. Table~\ref{table:fed_vs_dist} shows that instead of mimicking FedAvg's convergence rate, P2PL without max norm synchronization results in 65\% slower convergence.

As described in section \ref{section:p2pdl}, CFA~\cite{CFA} starts with a consensus step before the first training epoch, causing each device to start with the average of its neighbors' randomly initialized model parameters. For popular Gaussian, Xavier~\cite{glorot2010understanding}, Kaiming~\cite{he2015delving}, and PyTorch-default initialization schemes, averaged random parameters concentrate around zero (the origin) instead of the empirically beneficial regions identified by these schemes. This greatly slows convergence in practice and produces 206\% slower convergence than FedAvg in  Table~\ref{table:fed_vs_dist}. For a fairer comparison between P2PL and CFA, we modify CFA's training step to use momentum updates identical to Alg.~\ref{alg:P2PL} and change its consensus step to use mixing weights identical to \eqref{eq:mixing-weights} (using device-specific consensus step $\forall t\; \epsilon^{(t)}_k = \frac{\sum_{j\in\mathcal{N}(k)} n_j}{n_k + \sum_{j\in\mathcal{N}(k)} n_j}$).
Even after these improvements, Table~\ref{table:fed_vs_dist} shows that modifying CFA still results in 75\% slower convergence than FedAvg. 

P2PL's max norm synchronization phase synchronizes all devices' model parameters only once before the first training phase and ensures that the synchronized parameters lie in the regions targeted by popular single-model initialization schemes. This increases the effectiveness of both training and consensus phase updates. P2PL replicates FedAvg's convergence with only one synchronization at the start of training.
\subsection{Test Performance Robust to Diverse Communication Graphs} \label{sec:graph_type_test}
\begin{table*}[t]
\centering
\footnotesize
\caption{\fontsize{9}{10}\selectfont
Entries indicate the first communication round when minimum test accuracy of $K{=}100$ devices crossed the $97\%$ test accuracy threshold. Centralized Mini-batch GD crossed the threshold in $3$ training epochs. *Our peer-to-peer algorithm.}
\label{table:mh_nonIID}
\begin{tabular}{@{}lccccc@{}}
\toprule
& \multicolumn{3}{c}{IID Data} & \multicolumn{2}{c}{Non-IID Data}\\ 
\midrule
 &  \multicolumn{2}{c}{Dataset Size Weights} &  Metropolis-Hastings Weights & \multicolumn{2}{c}{Dataset Size Weights}\\ 
\midrule
\multirow{2}{*}{Federated Setting} & FedAvg & SCAFFOLD & FedAvg & FedAvg & SCAFFOLD\\
\cmidrule(lr){2-3}  \cmidrule(lr){4-4} \cmidrule(l){5-6}
 & 99  & 117 & 91 & 525 &  434\\
\midrule
Communication Graph & P2PL* & DSGD & P2PL* & P2PL* & DSGD\\
\cmidrule(r){1-1} \cmidrule(lr){2-3}  \cmidrule(lr){4-4} \cmidrule(l){5-6} 
Complete  & 87 & 5459 & 95 & 493 & 5519\\
2D Square Grid & 225 & 6719 & 239 & 5327& 15179\\
Erd\H os-R\'enyi & 220 & 7139 & 255 & 7135 & 17639\\
Watts-Strogatz & 284 & 6839 & 315 & 7196 & 16119\\
Star & 230 & 9179 & - & 7657 & 34139\\
3D Random Geometric & 769 & 7379 & 1479 & 9138 & 24599\\
Random Tree & 1278 &7979 &1874 & 9843 & 44699\\
Cycle & 2109 & 7619 & 2291 & 9894 & 41879\\
\bottomrule
\end{tabular}
\label{table2}
\end{table*}

\begin{figure}[t]
    \fontsize{9}{10}\selectfont
    \centering
    \includegraphics[width=0.8\columnwidth]{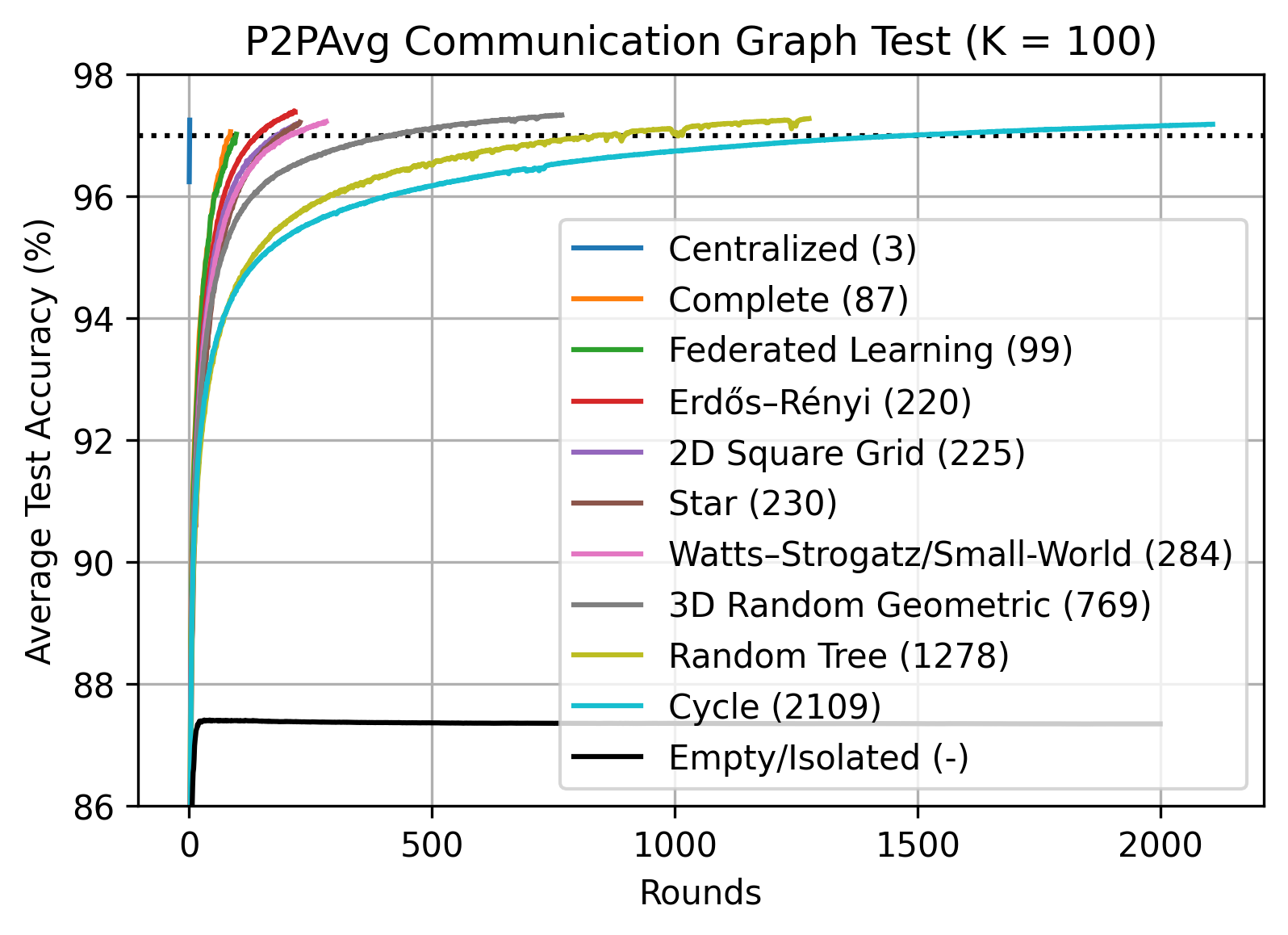}
    \caption{\fontsize{9}{10}\selectfont
    \textbf{P2PL convergence on various communication graphs} ($K{=}100$). Average test accuracy vs. rounds. Rounds to convergence in parentheses. Avg. accuracy may exceed $97\%$ before min. accuracy.}
    \label{fig:graph_type}
\end{figure}

We empirically show that P2PL, a peer-to-peer framework, enables all devices in diverse communication graphs to achieve test performance matching federated or centralized frameworks. The communication graph only impacts the rate of convergence. To show this for IoT environments, we consider several network topologies satisfying assumptions in Sec.~\ref{sec:comms}: complete, star\footnote{The distributed star graph has 1 central device and 99 peripheral devices, while federated settings have a server in the center and 100 peripheral devices.}, cycle, 2D square grid, and empty/isolated graphs along with 4 types of random graphs:

\begin{enumerate}[nolistsep, leftmargin=0.5cm]
    \item 3D Random Geometric Graphs (RGGs)~\cite{RGG}: device communication radius $r$ is selected to reflect ad hoc wireless network topologies with average degree of $4$. Average clustering coefficient: $0.478$; average shortest path length: $8.625$.
    
    \item Erd\H os-R\'enyi Graphs~\cite{erdhos1959random}: $p$, probability of an edge, is selected such that our graphs are connected and have an average degree of $4.653$. They have smaller average clustering coefficient $(0.025)$ and smaller average shortest path length $(3.552)$ than 3D RGGs.
    
    \item Watts-Strogatz (Small-World) Graphs~\cite{watts1998collective}: each device connects to 4 neighbors and the rewiring probability $p$ is selected to produce higher average clustering coefficient ($0.422$) and larger average path length ($5.885$) than our Erd\H os-R\'enyi graphs.
    
     \item Random Tree Graphs: we sample uniformly at random form all trees graphs on 100 nodes. They have the highest average shortest path length (10.922) and an average clustering coefficient of 0.
\end{enumerate}
Fig.~\ref{fig:graph_type} demonstrates that P2PL does not incur any test performance penalty when applied to large and diverse network topologies -- using a peer-to-peer framework only incurs a convergence rate penalty. The empty graph's flatlining ${\approx}87\%$ average accuracy highlights the importance of inter-device communication and the averaging consensus step \eqref{eq:consensus}.
Graphs with large average shortest path lengths (cycle: $25.253$; tree: $10.922$; 3D RGG: $8.625$) have slower information diffusion during P2PL's consensus phase and converge more slowly.

Table~\ref{table2} shows a comparison to Distributed Stochastic Gradient Descent (DSGD), which requires significantly more communication rounds than P2PL since DSGD performs one consensus step after each local mini-batch gradient update. Furthermore, results in Table~\ref{table2} show P2PL with a complete graph outperforming FedAvg and SCAFFOLD federated algorithms due to the max norm synchronization phase.

\subsection{Doubly Stochastic Mixing Weights: Metropolis-Hastings} \label{sec:mh}
P2PL exhibits convergence even with row stochastic mixing weights~\eqref{eq:consensus}. In comparison, apart from a 4-device experiment in~\cite{CFA}, all convergence analyses and experiments in related works~\cite{jiang2017collaborative, lian2017can, cooperativeSGD, pmlr-v119-koloskova20a, li2021communicationefficient, CFA} assume a 
symmetric doubly stochastic mixing matrix. This means that each pair of devices must assign the same importance to each other's parameters during consensus updates -- regardless of disparate local dataset sizes and local network topologies.

We test P2PL with doubly stochastic Metropolis-Hastings (M-H) mixing weights: neighbors $k$ and $i$ assign weight $\frac{1}{1+\max\{|\mathcal{N}(k)|, |\mathcal{N}(i)|\}}$ to each others' parameters; to ensure that the consensus step is a convex combination -- a device's own parameters receive the remaining weight. M-H weights produce symmetric doubly stochastic mixing weights for all undirected graphs~\cite{gossip-metropolis-hastings-mixing}. Table~\ref{table:mh_nonIID} shows that P2PL with M-H weights converges for all graphs except the star graph, which exceeds our communication budget and makes extremely slow but steady progress due to small mixing weights ($0.01$). The overall results validate previous theoretical convergence analysis on diverse communication graphs. However, they show that dataset size-dependent mixing weights can produce significant speedup. Thus, the results support using singly stochastic mixing weights in practice.


\subsection{Pathological Non-IID Data Distribution}
\label{sec:nonIID}
In practice, local training datasets at each device may not be independently and identically distributed (IID): $\mathcal{D}_k$ may not be representative of $\mathcal{D}$, the full training dataset at the global level. Even when most devices' training data corresponds to only 2 digits, Table~\ref{table2} shows that P2PL outperforms DSGD and converges on all tested communication graphs in a number of rounds on the same order of magnitude as federated learning with 1\% participation~\cite{pmlr-v54-mcmahan17a}.

\subsection{Link Failure Resiliency and Sparse Gossip Communication} 
\label{sec:gossip}
\begin{table}[t]
\footnotesize
\centering
\caption{\fontsize{9}{10}\selectfont
P2PL with $K{=}100$ Erd\H os-R\'enyi communication graph and link failures. Convergence: minimum test accuracy crosses $97\%$.}
\begin{tabular}{lcccc}
\toprule
 $1-C$: Prob. of Link Failure & 0 & 50\% & 75\% & 87.5\%\\
 \midrule
 Rounds to Convergence & 220 & 306 & 515 & 945\\
\bottomrule
\end{tabular}
\label{table:comm_sparsity}
\end{table}
Wireless device-to-device links in urban IoT environments suffer from fading and shadowing that lead to unrecoverable transmissions~\cite{walrand2017communication}. We model this with $C$ as the probability of a successful device-to-device transmission. For each directional communication between neighbors, we independently drop (ignore) transmitted parameters with probability $1-C$. This corresponds to a dynamic communication graph: a random directed subgraph of the underlying static undirected communication graph is active in each consensus step. Table~\ref{table:comm_sparsity} shows that all devices  in an underlying Erd\H os-R\'enyi communication graph achieve $97\%$ test accuracy and that P2PL is resilient to link failures.
This model for link failures is also equivalent to a sparsified random gossip communication model with ideal links: devices send parameters to each neighbor independently with probability $1-C$. Since a logarithmic decrease in $C$ results in a subexponential increase convergence rounds, a random gossip protocol can be used to improve P2PL's communication efficiency. 

\section{Conclusion}

As beyond-5G IoT environments rapidly evolve, peer-to-peer deep learning approaches are driving new advancements by solving federated learning's scaling concerns. P2PL is a practical and privacy-preserving device-to-device deep learning framework that is resilient to real-world conditions and enables large, diverse networks of devices to collaboratively achieve test performance on par with federated and centralized learning.





\bibliographystyle{IEEEbib}

\bibliography{bibfile}

\end{document}